\documentclass[runningheads]{llncs}

\usepackage{graphicx}
\usepackage[english]{babel}
\usepackage{multirow}
\usepackage{hyperref}
\usepackage{ amssymb }
\usepackage{ amsmath}
\usepackage{caption}
\usepackage{csquotes}
\usepackage{subcaption}
\usepackage{setspace}
\usepackage{pdfpages}
\usepackage[font=small,labelfont=bf]{caption}

\def\mi#1{\mathit{#1}}
\def\mc#1{\mathcal{#1}}
\newlength{\Oldarrayrulewidth}

\addto\extrasenglish{%
}
\begin{document}
\title{A Framework for Extracting and Encoding Features from Object-Centric Event Data}

\author{Jan Niklas Adams\inst{1} \and
Gyunam Park\inst{1} \and
Sergej Levich\inst{3} \and 
Daniel Schuster\inst{2,1}  \and 
Wil M.P. van der Aalst\inst{1,2}}
\authorrunning{J. N. Adams et al.}

\institute{Process and Data Science, RWTH Aachen University, Aachen, Germany \\ \email{\{niklas.adams,gnpark,schuster,wvdaalst\}@pads.rwth-aachen.de} \and Fraunhofer Institute for Applied Information Technology \and
Information Systems Research, University of Freiburg, Freiburg, Germany\\
\email{sergej.levich@is.uni-freiburg.de}}
\titlerunning{Extracting and Encoding Features from Object-Centric Event Data}
\maketitle              
\setcounter{footnote}{0}
\begin{abstract}
Traditional process mining techniques take event data as input where each event is associated with exactly one object. An object represents the instantiation of a process.
Object-centric event data contain events associated with multiple objects expressing the interaction of multiple processes. As traditional process mining techniques assume events associated with exactly one object, these techniques cannot be applied to object-centric event data.
To use traditional process mining techniques, the object-centric event data are flattened by removing all object references but one. The flattening process is lossy, leading to inaccurate features extracted from flattened data. Furthermore, the graph-like structure of object-centric event data is lost when flattening. In this paper, we introduce a general framework for extracting and encoding features from object-centric event data. We calculate features natively on the object-centric event data, leading to accurate measures. Furthermore, we provide three encodings for these features: \textit{tabular}, \textit{sequential}, and \textit{graph-based}. While tabular and sequential encodings have been heavily used in process mining, the graph-based encoding is a new technique preserving the structure of the object-centric event data. We provide six use cases: a visualization and a prediction use case for each of the three encodings. We use explainable AI in the prediction use cases to show the utility of both the object-centric features and the structure of the sequential and graph-based encoding for a predictive model.

\end{abstract}
\section{Introduction}
\label{sec:introduction}
{
\begin{figure}
    \centering
    \includegraphics[width=\columnwidth]{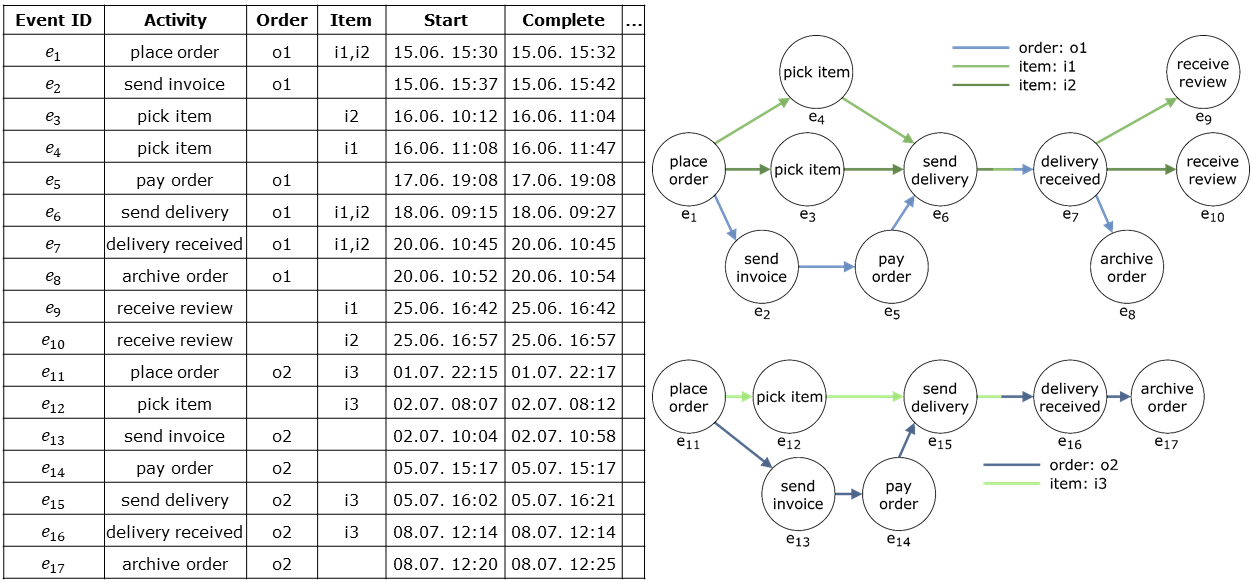}
    \caption{An object-centric event log and the underlying structure of events. The left-hand side depicts the event log. Events may be associated with multiple objects of different object types (here: Order and Item). The right-hand side shows the graph of directly-follows relationships for the events given by the objects. 
    An event with multiple objects may have multiple predecessor events.
    }
    \label{fig:OCEL_example}
\end{figure}}
\textit{Process mining}~\cite{ProcessMiningDSIA} is a branch of computer science producing data-driven insights and actions from event data generated by processes. These insights are typically grouped into three categories: process discovery, conformance checking, and enhancement. Process discovery techniques create process models describing the possible paths of actions in a process. Conformance checking techniques quantify and qualify the correspondence between a process model and event data. Process enhancement techniques take an encoding of features of the event data as input and deliver insights, predictions, or actions as output. Such enhancement techniques include process performance analysis~\cite{ChiaoPerformanceAnalysis,predictionservicetimeICSOC}, prediction~\cite{ImpactContextOutcome,LSTMPrediction,IntervalBaseTimePrediction} or clustering of similar process executions~\cite{ClusteringSubModels}.

Generally, process enhancement techniques encode features of event data in either of two ways: as a table \cite{GeneralFrameworkCorrelating,DBLP:conf/otm/DongenCA08} or as a set of sequences~\cite{DBLP:journals/dss/EvermannRF17,LSTMPrediction,DBLP:conf/bpm/LeontjevaCFDM15}. In a tabular encoding, each row corresponds to feature values for, e.g., an event.
This tabular encoding is used, for example, for regression, decision trees, and feed-forward neural networks. However, each process execution (also: case) is a timely ordered sequence of events. Therefore, summarizing event data to tabular encoding removes the sequential structure of the event data. Since this structure itself is meaningful, sequential encodings were developed~\cite{DBLP:conf/bpm/LeontjevaCFDM15}. These encodings represent each process execution as a sequence of feature values and are used for predictive models considering sequentially encoded data, such as LSTMs~\cite{LSTMPrediction}, or to visualize the variant of the process execution.

Traditional process mining builds on two central assumptions: Each event is associated with exactly one \textit{object} (the case) and each object is of the same type. Each object is associated with a sequence of events. A traditional \textit{event log}, therefore, describes a collection of homogeneously typed, isolated event sequences. 
This is a valid assumption when analyzing, e.g., the handling of insurance claims. In this example, each object describes an instantiation of the same type: an insurance claim. Events are associated to exactly one insurance claim.
However, real-life information systems often paint another picture: Events may be related to multiple objects of different types~\cite{DiscoveringObjectCentricPetriNets,PrecisionFitnessOCPM,GraphDatabases,CausalProcessMining}. The most prominent example of information systems generating event data with multiple associated objects are ERP systems. Objects in such systems would correspond to, e.g., an order, different items of this order, and invoices in an order-to-cash process. Consider the simplified example of an order handling process depicted in \autoref{fig:OCEL_example}. An event may be related to objects of type order, item, or both. 
An event with multiple objects may have multiple predecessor events.
Therefore, the structure of an \textit{Object-Centric Event Log (OCEL)} resembles a graph, not a sequential structure as is assumed in traditional process mining.

{
\begin{figure}[t]
\centering
\begin{subfigure}{\textwidth}
        \includegraphics[width=\columnwidth]{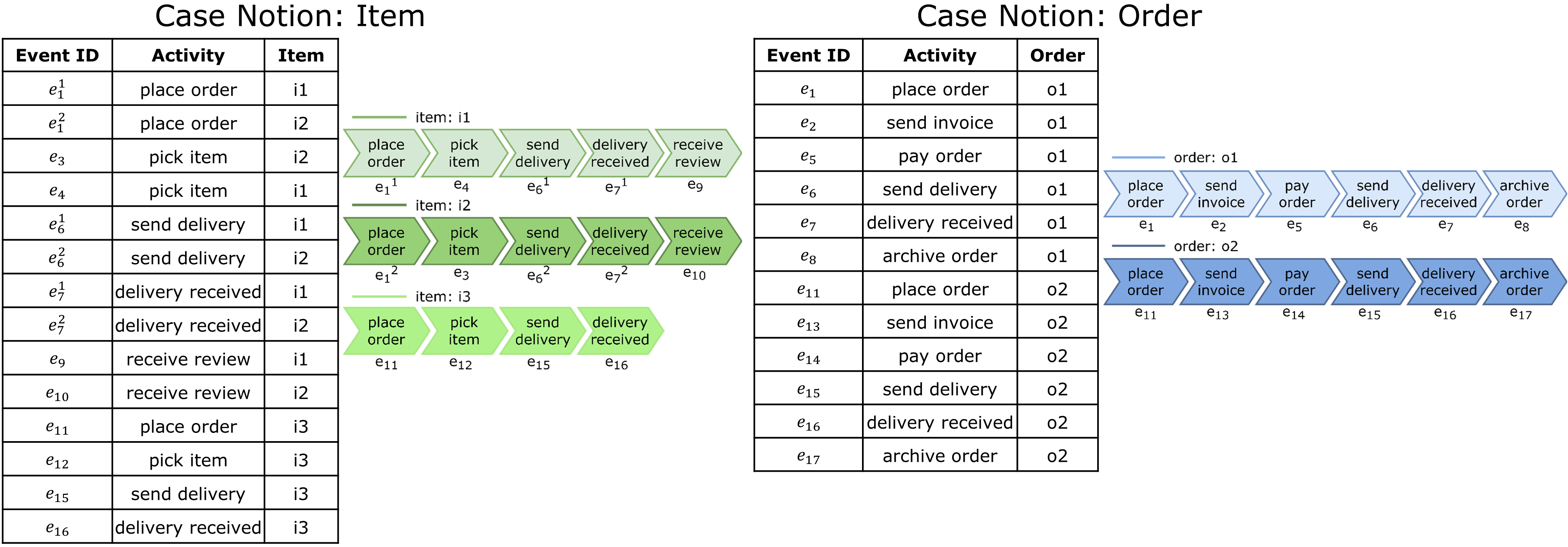}
    \end{subfigure}
    \begin{subfigure}{\textwidth}
    \centering
        \includegraphics[width=0.89\columnwidth]{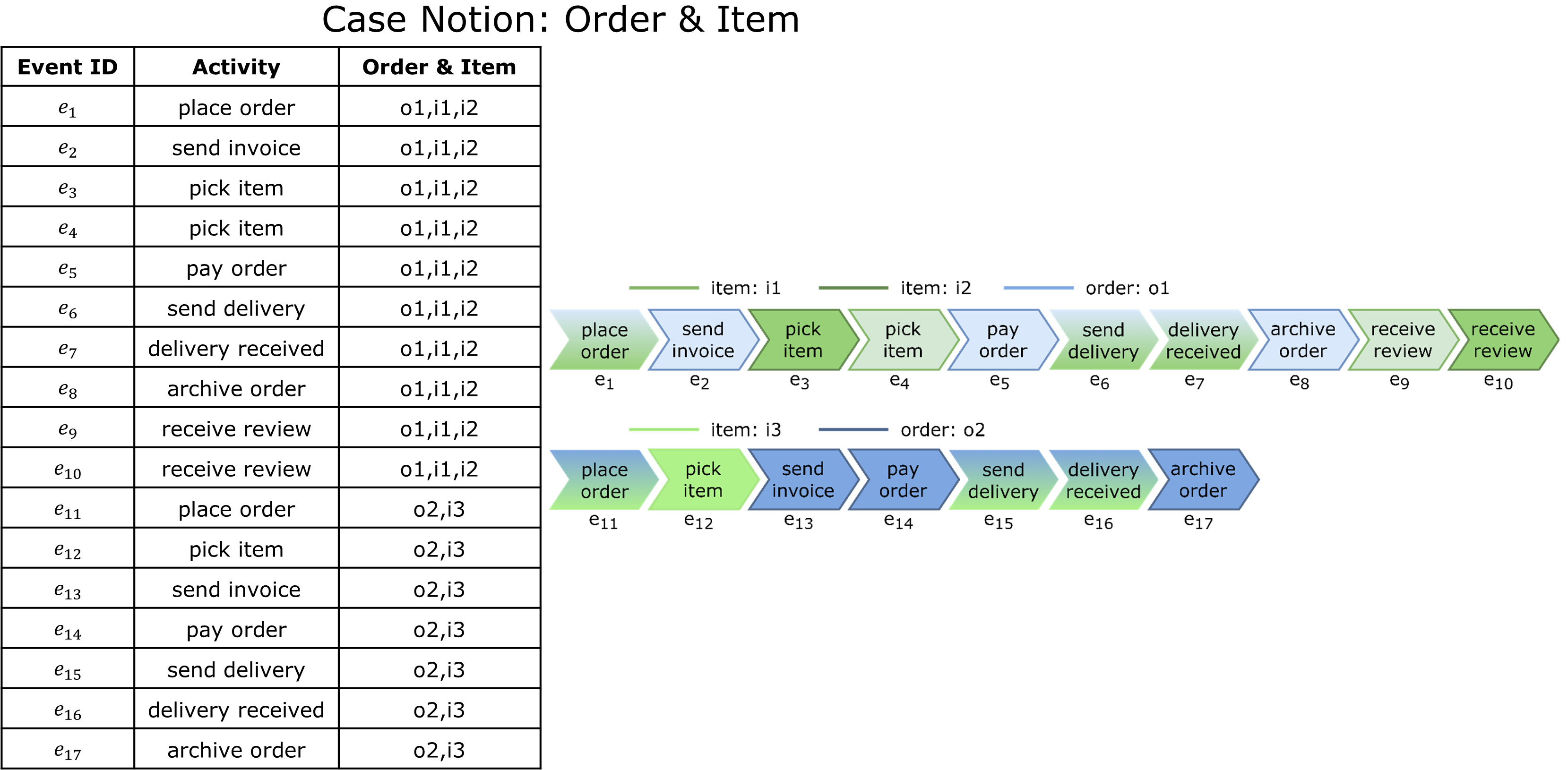}
    \end{subfigure}
    \caption{Flattening an object-centric event log (cf. \autoref{fig:OCEL_example}) such that it can be used for traditional process enhancement techniques. The event log is transformed into sequences of a chosen case notion. Due to deficiency, convergence and divergence, the features calculated on a flattened log might be misleading, e.g., through missing events. Furthermore, the graph-like structure of the original event log is lost.
    }
    \label{fig:flatteing}
\end{figure}}

This gap between OCELs and traditional process enhancement techniques is currently bridged by \textit{flattening} an event log~\cite{OCPMDivergenceAndconvergence}, i.e., mapping an OCEL into traditional event log format by enforcing a homogeneous, sequential structure.
This involves two steps: Choosing a \textit{case notion} and duplicating events with multiple objects of that notion. All objects not included in this case notion are discarded. Flattening the event log of \autoref{fig:OCEL_example} is depicted in \autoref{fig:flatteing} for three different case notions. The first two are case notions of a single object type~\cite{OCPMDivergenceAndconvergence}. 
The third case notion is a composite case notion of co-appearing orders and items, i.e., an order and all corresponding items. 
The flattened event data may be used as input for traditional process enhancement techniques. 

However, flattening manipulates the information of the object-centric event log. The problems related to flattening are \textit{deficiency} (disappearing events)~\cite{DiscoveringObjectCentricPetriNets},  \textit{convergence} (duplicated events)~\cite{OCPMDivergenceAndconvergence} and \textit{divergence} (misleading directly-follows relations)~\cite{CausalProcessMining,OCPMDivergenceAndconvergence}.
We showcase divergence using an example. One might use a composite case notion of order and item to flatten the event log (cf. \autoref{fig:flatteing} Case Notion: Order \& Item). All orders and items related through events form one composite object, i.e., o1,i1,i2, and o2,i3. The events of these objects are flattened to one sequence, introducing inaccurate precedence constraints.
E.g., events $e_3$ and $e_4$, which describe an item being picked, are now sequentially ordered, indicating some order between them. However, the original event data show that these two picking events are independent. The same holds for the relationship of pick item and pay order: The object-centric event data do not indicate any precedence constraint. However, the sequential representation enforces one.

{
\begin{figure}[t]
    \centering
    \includegraphics[width=0.90\columnwidth]{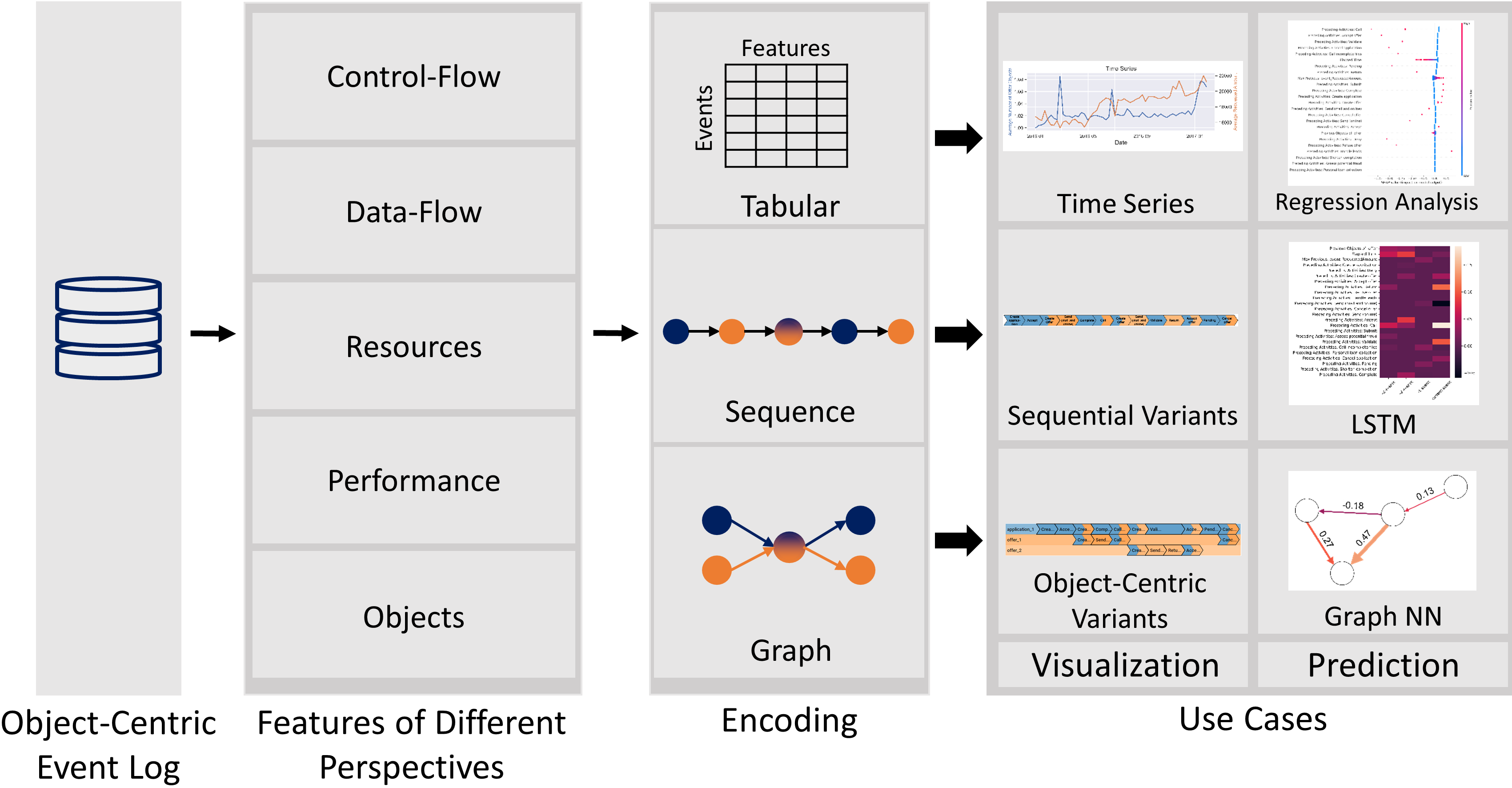}
    \caption{Our framework enables accurate feature extraction for object-centric event data. Furthermore, we provide three encodings for object-centric features: tabular, sequential, and graph-based. We present a visualization and prediction use case for each encoding.}
    \label{fig:overview}
\end{figure}}

The three problems of flattening have major consequences on the quality of the calculated features of the flattened OCEL: Due to missing events, duplicated events, or wrong precedence constraints, many features deliver incorrect results (cf. \autoref{sec:features}).  
Furthermore, the tabular or sequential encoding constructed from these features does not preserve the graph-like structure of the event data, removing important structural information. Therefore, features for OCELs can not accurately be extracted and encoded.

To solve the previously mentioned problem, an approach is necessary that calculates features natively on the object-centric event data and enables a graph-based encoding preserving the actual structure of the event log. In this paper, we introduce a general framework for extracting and encoding features for object-centric event data (cf. \autoref{fig:overview}), providing two contributions:
\textbf{1)} We translate the computation of the features introduced in the framework of de Leoni et. al \cite{GeneralFrameworkCorrelating} to the object-centric setting, providing accurate measures. \textbf{2)} We provide three different encodings to represent the extracted features for different algorithms and methods: tabular, sequential, and graph-based. Using features and encoding, we provide six use cases.
These use cases showcase the generalizability of our framework to a plethora of different tasks. 
We use one visualization and one prediction use case for each encoding.
In the prediction use cases, we depict how the different features and the structure of the encodings are utilized by predictive models, leveraging on explainable AI and \textsc{SHAP} values~\cite{shap}. These contributions may be used as a foundation for new algorithms, new visualizations, new machine learning models, more accurate predictions, and more.

This paper is structured as follows. First, we discuss related work on feature extraction and encoding in \autoref{sec:relwork}. We introduce object-centric event data and process executions in \autoref{sec:prelim}. In \autoref{sec:features}, we provide an overview of native feature calculation for object-centric event data. In \autoref{sec:encodings}, we define three encodings for object-centric features. \autoref{sec:casestudies} depicts our six use cases for features and their encodings. We conclude this paper in \autoref{sec:conclusion}.

\section{Related Work}
\label{sec:relwork}
\begin{table}[t]
\centering
\caption{Process enhancement techniques and supporting frameworks.}
\label{tab:related-work}
\resizebox{\textwidth}{!}{%
\begin{tabular}{|cc|cc|ccc|c|}
\hline
\multicolumn{2}{|c|}{\multirow{2}{*}{}} &
  \multicolumn{2}{c|}{Feature Extraction} &
  \multicolumn{3}{c|}{Feature Encoding} &
  \multirow{2}{*}{\begin{tabular}[c]{@{}c@{}}Existing\\ work\end{tabular}} \\ \cline{3-7}
\multicolumn{2}{|c|}{} &
  \multicolumn{1}{c|}{Object-centric} &
  Flattened &
  \multicolumn{1}{c|}{Tabular} &
  \multicolumn{1}{c|}{Sequential} &
  Graph-based &
   \\ \hline
\multicolumn{1}{|c|}{\multirow{3}{*}{\begin{tabular}[c]{@{}c@{}}(a) Process\\ enhancement\\ techniques\end{tabular}}} &
  \textit{P1} &
  \multicolumn{1}{c|}{} &
  \checkmark &
  \multicolumn{1}{c|}{\checkmark} &
  \multicolumn{1}{c|}{} &
   &
  \cite{DBLP:conf/otm/DongenCA08,DBLP:conf/otm/FolinoGP12} \\ \cline{2-8} 
\multicolumn{1}{|c|}{} &
  \textit{P2} &
  \multicolumn{1}{c|}{} &
  \checkmark &
  \multicolumn{1}{c|}{} &
  \multicolumn{1}{c|}{\checkmark} &
   &
  \cite{DBLP:conf/bpm/LeontjevaCFDM15,DBLP:journals/dss/EvermannRF17,LSTMPrediction} \\ \cline{2-8} 
\multicolumn{1}{|c|}{} &
  \textit{P3} &
  \multicolumn{1}{c|}{} &
  \checkmark &
  \multicolumn{1}{c|}{} &
  \multicolumn{1}{c|}{} &
  \checkmark &
  \cite{9004296,9533742,doi:10.1080/12460125.2020.1780780} \\ \hline
\multicolumn{1}{|c|}{\multirow{3}{*}{\begin{tabular}[c]{@{}c@{}}(b) \\ Frameworks\end{tabular}}} &
  \textit{F1} &
  \multicolumn{1}{c|}{} &
  \checkmark &
  \multicolumn{1}{c|}{\checkmark} &
  \multicolumn{1}{c|}{} &
   &
  \cite{GeneralFrameworkCorrelating} \\ \cline{2-8} 
\multicolumn{1}{|c|}{} &
  \textit{F2} &
  \multicolumn{1}{c|}{} &
  \checkmark &
  \multicolumn{1}{c|}{} &
  \multicolumn{1}{c|}{\checkmark} &
   &
  \cite{FrameworkSequential,FrameworkSequential2} \\ \cline{2-8} 
\multicolumn{1}{|c|}{} &
  \textbf{This paper} &
  \multicolumn{1}{c|}{\textbf{\checkmark}} &
  \textbf{} &
  \multicolumn{1}{c|}{\textbf{\checkmark}} &
  \multicolumn{1}{c|}{\textbf{\checkmark}} &
  \textbf{\checkmark} &
   \\ \hline
\end{tabular}%
}
\end{table}

A plethora of process enhancement techniques exist in the literature, including process performance analysis, predictive process monitoring, and trace clustering~\cite{ProcessMiningDSIA}.
Such techniques use encoded features extracted from an event log as input.
\autoref{tab:related-work}(a) shows three categories of techniques using different feature extraction (i.e., feature extractions using 1. OCELs and 2. flattened event logs) and encoding (i.e., 1. tabular, 2. sequential, and 3. graph encoding) approaches with representative examples.
First, \textit{P1} represents the techniques using features extracted from flattened event logs and encoded as tabular formats.
For instance, van Dongen et al.~\cite{DBLP:conf/otm/DongenCA08} use tabular encoding by transforming an event log into feature-outcome pairs to predict remaining times using non-parametric regression.
Also, in~\cite{DBLP:conf/otm/FolinoGP12}, an event log is encoded into a tabular format with additional features on context, e.g., resource availability, to predict processing times.
Second, techniques in \textit{P2} also use features based on flattened event logs but encoded as sequential formats.
Leontjeva et al.~\cite{DBLP:conf/bpm/LeontjevaCFDM15} propose \emph{complex sequence encoding} to encode an event log to sequences to predict the outcome of an ongoing case.
To predict the next activity of an ongoing case, Evermann et al.~\cite{DBLP:journals/dss/EvermannRF17} encode control-flow features using \emph{embedding} techniques, whereas Tax et al.~\cite{LSTMPrediction} use \emph{one-hot encoding}.
Finally, \textit{P3} consists of techniques using features extracted from flattened event logs and encoded as graph formats.
Philipp et al.~\cite{9004296} encode an event log to a graph where each node represents an activity, and each edge indicates the relationship between activities. 
The graph is used to learn a \emph{Graph Neural Network (GNN)} to predict process outcomes.
Venugopal et al.~\cite{9533742} extend \cite{9004296} by annotating nodes with temporal features. 
They use GNNs to predict the next activity and next timestamp of an event.
Instead of representing a node as an activity, Harl et al.~\cite{doi:10.1080/12460125.2020.1780780} uses one-hot encoding of an activity to represent a node to deploy \emph{gated graph neural network} that provides the explainability based on \emph{relevance score}. 

Furthermore, to support the development of process enhancement techniques using different feature extraction and encoding, several frameworks have been proposed (cf. \autoref{tab:related-work}(b)).
First, De Leoni~\cite{GeneralFrameworkCorrelating} in \textit{F1} suggest a framework to compute features using flattened event logs and encode them to tables.
Second, Becker et al.~\cite{FrameworkSequential} and Di Francescomarino et al.~\cite{FrameworkSequential2} in \textit{F2} propose frameworks for techniques for sequentially encoding extracted features.
To the best of our knowledge, no framework supporting graph encoding exists.

Despite the limitations of flattened event logs to extract misleading features, no study has been conducted to develop process enhancement techniques using features based on OCELs.
In this work, we provide a framework for extracting and encoding features based on OCELs, with the goal of facilitating the development of object-centric process enhancement approaches.
Our proposed framework supports all existing encoding formats, i.e., tabular, sequential, and graph, to be used for different algorithms and methods.

\section{Object-Centric Event Data}
\label{sec:prelim}
Given a set $X$, the powerset $\mc{P}(X)$ denotes the set of all possible subsets. A sequence $\sigma:\{1,\ldots,n\}\rightarrow X$ of length $\mi{len}(\sigma) = n$ assigns order to elements of $X$. We denote a sequence with $\sigma = \langle x_1, \ldots , x_n \rangle$ and the set of all sequences over $X$ with $X^*$. We overload the notion $x\in \sigma$ to express $x \in \mi{range}(\sigma)$.

A graph is a tuple $G = (V,E)$ of nodes $V$ and edges $E\subseteq V \times V$. The set of all subgraphs of $G$ is given by
$\mi{sub}(G) = \{(V', (V'\times V') \cap E ) \mid V' \subseteq V \}$.
A path connects two distinct nodes through edges. The set of paths between two nodes $v,v' \in V, v\neq v'$ is defined by  $\mi{path}_G(v,v') = \{\langle (v,v_1),(v_1,v_2),\ldots, (v_k,v') \rangle \in E^*\}$. 
Two distinct nodes are connected if the set of paths between them is not empty $\mi{path}_G(v,v') \neq \emptyset$. The distance between two nodes is the length of the shortest path $\mi{dist}_G(v,v') = len(\sigma_d)$ such that $\sigma_d \in \mi{path}_G(v,v') \wedge \neg \exists_{\sigma_{d}' \in \mi{path}_G(v,v') }\ \allowbreak len(\sigma_d) \allowbreak > len(\sigma_d') $.
A graph $G=(V,E)$ is connected iff a path exists between all edges $\forall_{v,v' \in V}\ v \neq v' \wedge \mi{path}_G(v,v') \neq \emptyset$. The set of connected subgraphs of $G=(V,E)$ is defined as follows $\mi{consub}(G)=\{G'\in \mi{sub}(G) \mid G' \text{ is connected}\}$.\

An event log is a collection of events associated with objects. Each event contains an activity, describing the executed action, a start and complete timestamp and additional attributes. Each object is associated to a sequence of events.
\begin{definition}[Event Log]
Let $\mathcal{E}$ be the universe of events, $\mc{O}$ be the universe of objects, $\mc{OT}$ be the universe of object types, $\mc{A}$ be the universe of activities, $\mc{C}$ be the universe of attributes and $\mc{V}$ be the universe of attribute values. Let $A \subseteq \mc{A}$ be a set of activities and $C \subseteq \mc{C}$ be a set of attributes. Each object is mapped to exactly one object type $\pi_{type}:\mc{O}\rightarrow \mc{OT}$.
An event log $L = (E,O,OT,\pi_\mi{ct},\pi_\mi{st},\pi_\mi{trace},\pi_\mi{act},\pi_\mi{att})$ is a tuple composed of 
\begin{itemize}
    \item[$\bullet$] events $E\subseteq \mathcal{E}$, objects $O\subseteq \mathcal{O}$, and object types $OT\subseteq \mathcal{OT}$,
    \item[$\bullet$] two time mappings for the completion $\pi_\mi{ct}:E\rightarrow \mathbb{R}$ and the start $\pi_\mi{st}:E\rightarrow \mathbb{R}$ of an event such that $\pi_\mi{st}(e)\leq \pi_\mi{ct}(e)$ for any $e\in E$,
    \item[$\bullet$] a mapping $\pi_\mi{trace}:O\rightarrow E^*$ mapping each object to a sequence of events such that $\forall_{o \in O}\ \pi_\mi{trace}(o) = \langle e_1,\ldots,e_n \rangle \wedge \forall_{i \in \{1,\ldots n-1\}}\ \pi_\mi{ct}(e_i)\leq \pi_\mi{ct}(e_{i+1})$,
    \item[$\bullet$] an activity mapping $\pi_\mi{act}:E\rightarrow A$ and,
    \item[$\bullet$] an attribute mapping $\pi_\mi{att}:E\times C\nrightarrow \mathcal{V}$.
\end{itemize}
\end{definition}
The table in \autoref{fig:OCEL_example} depicts an example of an OCEL. A row corresponds to one event. Sorting the events of an object in timely order, we retrieve the event sequence for the object, e.g., $\pi_\mi{trace}(\text{i3})=\langle \text{place order},\allowbreak \text{pick item},\allowbreak  \text{send delivery},\allowbreak \text{delivery received}\rangle$.
The relationships between objects can be expressed in the form of a graph, connecting objects that share events.
\begin{definition}[Object Graph]
Let $L = (E,O,OT,\pi_\mi{ct},\pi_\mi{st},\pi_\mi{trace},\pi_\mi{act},\pi_\mi{att})$ be an event log. We denote the objects of an event $e{\in} E$ with $\pi_\mi{obj}(e) {=} \{o {\in} O \mid e {\in} \pi_\mi{trace}(o)\}$. The object graph $\mi{OG}_L {=} (O,I)$ is an undirected graph of nodes $O$ and edges of object interactions $I {=} \{\{o,o'\}{\subseteq} O \mid o {\neq} o' \wedge \exists_{e\in E}\ \{o,o'\} \subseteq \pi_\mi{obj}(e)\}$.
\end{definition}
Objects which are directly or transitively connected in the object graph depend on each other by sharing events. In traditional process mining, a process execution (case) is the event sequence of one object. We use the definitions of process executions~\cite{AdamsDefiningCases} and generalize this notion such that a process execution is the set of events for multiple, connected objects.
\begin{definition}[Process Execution]
Let $L = (E,O,OT,\pi_\mi{ct},\pi_\mi{st},\pi_\mi{trace},\pi_\mi{act},\allowbreak \pi_\mi{att})$ be an event log and $\mi{OG}_L = (O,I)$ be the corresponding object graph. A process execution $p=(O',E')$ is a tuple of objects $O' \subseteq O$ and events $E'\subseteq E$ such that
    $e'\in E' \Leftrightarrow \pi_\mi{obj}(e')\subseteq O'$ and
    $O'$ forms a connected subgraph in $\mathit{OG}_L$.
\end{definition}
We define two techniques to extract process executions from an OCEL. These two techniques are two out of many possible process execution extraction techniques. 
The first technique extracts process executions based on the connected components of the object graph. All transitively connected objects form one process execution. This might lead to large executions for entangled event logs. Therefore, we introduce the leading type extraction. A process execution is constructed for each object of a chosen leading object type. Connected objects are added to this process execution unless a connected object of the same type has a lower distance to the leading object. This limits executions in size but also removes dependencies.
\begin{definition}[Execution Extraction]
Let $L {=} (E,O,OT,\pi_\mi{ct},\pi_\mi{st},\pi_\mi{trace},\allowbreak \pi_\mi{act},\allowbreak \pi_\mi{att})$  be an event log. An execution extraction $\mi{EX}{\subseteq}  \mi{consub}(\mi{OG}_L)$ retrieves connected subgraphs from the object graph. A subgraph $\mi{ex} = (O',I') \in \mi{EX}$ is mapped to a process execution 
$f^\mi{extract}(\mi{ex},L) = (O',E') $ with $E' = \{e\in E\mid O' \cap \mi{obj}(e) \neq \emptyset\} $. We define two extraction techniques:
\begin{itemize}
    \item[$\bullet$] $\mi{EX}_\mi{comp}(L) = \{G \in \mi{consub}(\mi{OG}_L) \mid \neg \exists_{G' \in \mi{consub}(\mi{OG}_L)}\ G \in \mathit{sub}(G')\}$, and
    \item[$\bullet$] $\mi{EX}_\mi{lead}(L,\mi{ot}) = \{ G \in \mi{lead\_graphs} = \{ G'=(O',I') \in \mi{consub}(OG_L) \mid  \exists_{o\in O'}\ \pi_\mi{type}(o) = ot \wedge \forall_{o'\in O'}\ \neg \exists_{o''\in O'}\ o'' \neq o' \wedge \pi_\mi{type}(o'') = \pi_\mi{type}(o') \wedge \mi{dist}_{G'}(o,o') > \mi{dist}_{G'}(o,o'')  \} \mid \neg \exists_{G'' \in \mi{lead\_graphs}} G \in \mi{sub}(G'')\}$ for $\mi{ot} \in OT$.
\end{itemize}
\end{definition}
When looking at the example of \autoref{fig:OCEL_example}, the process executions retrieved by applying $\mi{EX}_\mi{comp}$ would be based on the connected components of the object graph, i.e., $\{o1,i1,i2\}$ and $\{o2,i3\}$. Using the leading type order, we would retrieve the same executions. Using item as the leading type, we would retrieve $\{i1,o1\}$, $\{i2,o1\}$ and $\{i3,o2\}$. 
\section{Object-Centric Features}
\label{sec:features}
This section deals with the problem resulting from flattening OCELs to apply process enhancement techniques: Features are calculated on the manipulated, flattened event data. Therefore, they might be inaccurate. We propose an object-centric adaptation of the features introduced by the seminal machine learning framework of de Leoni et al.~\cite{GeneralFrameworkCorrelating}. We calculate them natively on the OCEL. Furthermore, we provide several new features recently introduced in the literature on object-centric process mining.
A feature is, generally, calculated for an event. It might describe a measure for the single event, in relationship to its process executions, or the whole system. 
\begin{definition}[Features]
Let $L = (E,O,OT,\pi_\mi{ct},\allowbreak \pi_\mi{st},\pi_\mi{trace},\pi_\mi{act},\pi_\mi{att})$ be an event log and $\mi{EX}{\subseteq}  \mi{consub}(\mi{OG}_L)$ be a set of extracted process executions. A feature 
$f_L{:}E {\times} \mi{EX}  {\nrightarrow} \mathbb{R}$ maps an event and a process execution onto a real number.
\end{definition}

{
\begin{figure}[t]
    \centering
    \includegraphics[width=\columnwidth]{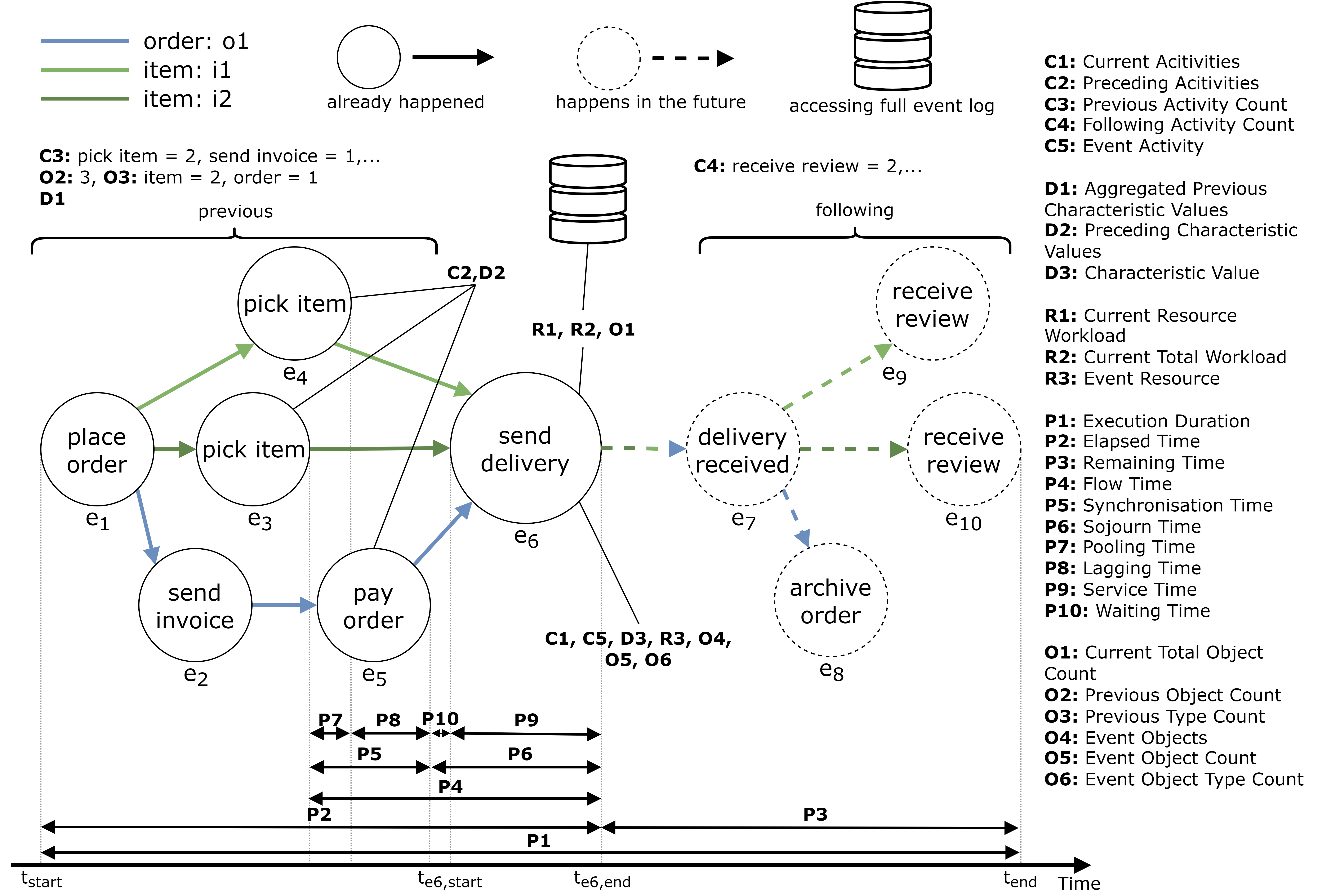}
    \caption{Overview of the features that can be extracted for event $e_6$. These features are the object-centric adaptations of~\cite{GeneralFrameworkCorrelating}.}
    \label{fig:feature_example}
\end{figure}}
The primary need for adapting traditional feature calculation arises from two main differentiations between object-centric and traditional event data: First, each event can have multiple predeccesors/successors, one for each object. Second, each event might have multiple objects of different types. The computation of features that are depended on previous and following behavior has to be adapted to the graph structure. The most obvious example are preceding activities: In traditional feature extraction, there is only one preceding activity for each event. In object-centric feature extraction, there are multiple preceding activities, one for each object. The graph-structure as well as the multiplicity of objects also enables the definition of new features leveraging on the graph structure and object (type) associations.
Previous (i.e., all events that happened before the considered event in an execution) and following events can be adapted in two ways: time-based (using the event's timestamp) and path-based (using path information of the graph). We use a simple time-based adaptation. However, the graph-based adaptation might give interesting new research directions. 

An overview of the features collected from an object-centric adaptation of de Leoni et al.'s framework~\cite{GeneralFrameworkCorrelating} and features recently introduced in the literature \cite{DiscoveringObjectCentricPetriNets,Opera} is depicted in \autoref{fig:feature_example}. Similar to de Leoni et al., we group features according to different perspectives: \textbf{C}ontrol-Flow, \textbf{D}ata-Flow, \textbf{R}esource, \textbf{P}erformance and \textbf{O}bjects. We, now, discuss the different perspectives and the adaptations that are necessary to apply them to the object-centric setting. \autoref{tab:features_overview} provides a qualitative evaluation of the impact of flattening on the resulting feature value: Features can be equal, they can be misleading/incorrect after flattening, and not be available for flat event data.
{
\setlength{\tabcolsep}{4pt}
\renewcommand{\arraystretch}{1.13}
\begin{table}[t]
    \centering
    \caption{Impact of flattening on calculated feature values. Calculating a feature for an event on object-centric vs. flattened data can lead to correct or misleading results. Some features only exist on object-centric event data. Most features are misleading due to the graph structure and object multiplicity.
    }
     \resizebox{0.98\textwidth}{!}{
    \begin{tabular}{|l|l|c|c|c|}\hline
   \multicolumn{2}{|c|}{\multirow{2}{*}{\textbf{Features}}} & \multicolumn{2}{|c|}{\textbf{Impact of flattening}}&\multirow{2}{*}{\textbf{Only available for OCEL}}\\\cline{3-4}
    \multicolumn{2}{|c|}{}&Correct&Misleading& \\\hline
    \multirow{2}{*}{Control-Flow}&\textbf{C1}, \textbf{C2}, \textbf{C3}, \textbf{C4}&&\checkmark&\\\cline{2-5}
    &\textbf{C5}&\checkmark&&\\\hline
    \multirow{2}{*}{Data}&\textbf{D1}, \textbf{D2}&&\checkmark&\\\cline{2-5}
    &\textbf{D3}&\checkmark&&\\\hline
    \multirow{2}{*}{Resource}&\textbf{R1}&&\checkmark&\\\cline{2-5}
    &\textbf{R2}, \textbf{R3}&\checkmark&&\\\hline
    \multirow{3}{*}{Performance}&\textbf{P1}, \textbf{P2}, \textbf{P3}, \textbf{P6}, \textbf{P10}&&\checkmark&\\\cline{2-5}
    &\textbf{P4}, \textbf{P5}, \textbf{P7}, \textbf{P8}&&&\checkmark\\\cline{2-5}
    &\textbf{P9}&\checkmark&&\\\hline
    \multirow{1}{*}{Objects}&\textbf{O1}, \textbf{O2}, \textbf{O3}, \textbf{O4},  \textbf{O5}&&&\checkmark\\\hline

    \end{tabular}}
    
    \label{tab:features_overview}
\end{table}
}

The main adaptations of the control-flow perspectives are concerned with the switch from sequential to graph-like control-flow. Multiple preceding activities (\textbf{C2}) as well as multiple current activities (\textbf{C1}) (endpoints of the current execution graph) are possible. For previous and following activities (\textbf{C3}, \textbf{C4}), we use a simple time-based adaptation.

The data-flow perspective needs slight adaptations for preceding characteristic values (\textbf{D2}). Since there might be multiple preceding values, these need to be aggregated. Previous characteristic values (\textbf{D1}) are adapted on a time basis, and the characteristic value (\textbf{D3}) needs no adaptation.

The resources perspective's features are mainly concerned with system-wide measurements, such as the workload of the current resource (\textbf{C1}) or the total system workload (\textbf{C2}). Therefore, this perspective remains mostly unaffected by a move to object-centricity. Future research might investigate new features derived from resource multiplicity per event. 

The performance perspective has recently been studied for new object-centric features~\cite{Opera}. Due to an event having multiple predecessors, the established performance measures can be extended by several features expressing the time for synchronization between objects (\textbf{P5}), the pooling time of an object type (\textbf{P7}), or the lag between object types before the event (\textbf{P8}). 

Finally, a new feature perspective concerning objects opens up. The paper introducing the discovery of object-centric Petri nets \cite{DiscoveringObjectCentricPetriNets} introduces some basic features of the object perspective. For example, an event's number of objects (\textbf{O5}), the event's number of objects of a specific type (\textbf{O6}), or the current system's total object count (\textbf{O1}). Investigations of additional features in this perspective, e.g., quantifying the relationships between objects through graph metrics on the object graph, might also be an interesting research direction.

\section{Feature Encodings}
\label{sec:encodings}
In this section, we tackle the absence of feature encodings that represent the graph-like structure of object-centric event data. We extend the currently used tabular and sequential encodings with a graph-based one and introduce all three encodings formally. Together with the formal definition of each encoding, we provide some common use cases, advantages, disadvantages and a continuation of our running example from \autoref{fig:OCEL_example}. As an example of extracted features we choose the number of previous objects (\textbf{O2}), the synchronization time (\textbf{P5}) and the remaining time (\textbf{P3}). The execution extraction for our example is the connected components extraction $\mi{EX}_\mi{comp}$. A tabular encoding is a common representation of data points used for many use cases, such as regression analysis, clustering, different data mining tasks, etc. 
{
\begin{figure}[t]
    \centering
    \includegraphics[width=0.95\columnwidth]{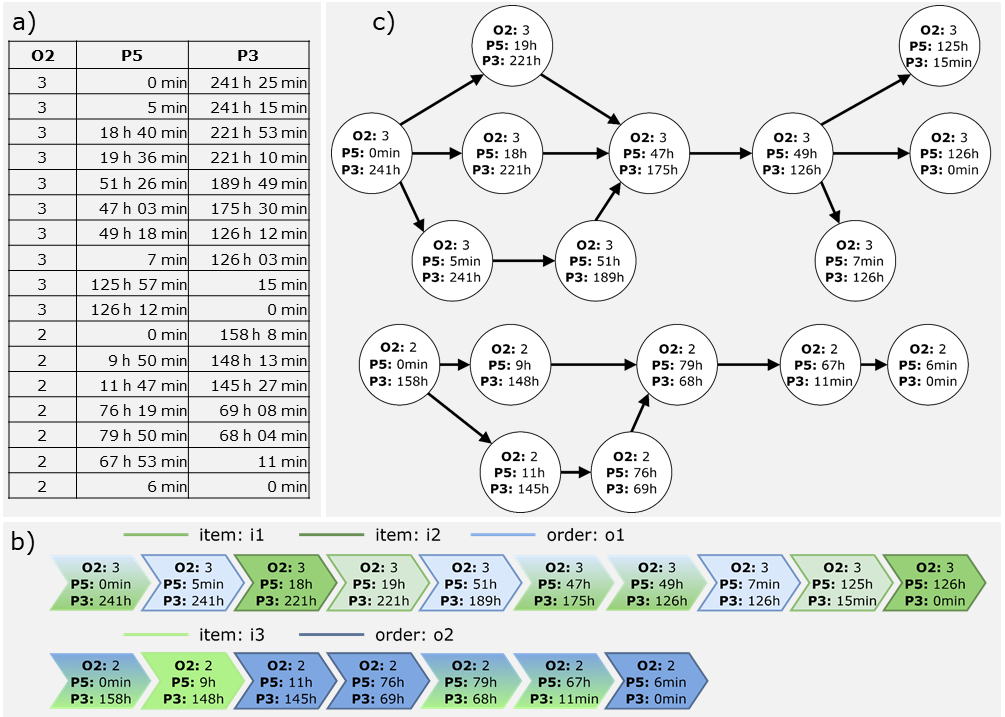}
    \caption{Example of tabular a), sequential b) and graph-based c) feature encodings for the running example in \autoref{fig:OCEL_example}. The graph-based encoding preserves the structural information from the OCEL.}
    \label{fig:encoding}
\end{figure}}
\begin{definition}[Tabular Encoding]
Let $L = (E,O,OT,\pi_\mi{ct},\pi_\mi{st},\pi_\mi{trace},\pi_\mi{act}, \allowbreak \pi_\mi{att})$ be an event log and $\mi{EX}\subseteq  \mi{consub}(\mi{OG}_L)$ be a set of process executions. Let $F_L\subseteq E \times \mi{EX} \nrightarrow \mathbb{R}$ be a set of features. The event feature table is defined by $\mi{tab}(e,f_L) = f_L(e,\mi{ex})$ for all $e\in E$, $\mi{ex}\in \mi{EX}$ (rows) and all $f_L \in F_L$ (columns).
\end{definition}
We depict an example of tabular encoding in \autoref{fig:encoding} a). Such an encoding is easily readable and versatile usable, however, the structural order information of the event log is lost in the process of tabular encoding. 
A sequential encoding is commonly used in sequence visualization, clustering, classification or next value predictions (cf. \autoref{sec:relwork}).
\begin{definition}[Sequential Encoding]
Let $L = (E,O,OT,\pi_\mi{ct},\pi_\mi{st},\pi_\mi{trace},\pi_\mi{act}, \allowbreak \pi_\mi{att})$ be an event log and $\mi{EX}\subseteq  \mi{consub}(\mi{OG}_L)$ be a set of extracted process executions. Let $F_L {=} \{f_{L,1}, \ldots, f_{L,m}\} {\subseteq} E {\times} \mi{EX} {\nrightarrow} \mathbb{R}$ be a set of features.
The sequential encoding of an execution  $\mi{ex}\in \mi{EX}$ is defined by $\mi{seq}(\mi{ex},F_L) = \langle (f_{L,1}(e_1,\mi{ex}),\allowbreak \ldots,\allowbreak f_{L,m}(e_1,\mi{ex})), \ldots ,\allowbreak (f_{L,1}(e_n,\mi{ex}),\allowbreak \ldots, \allowbreak f_{L,m}(e_n,\mi{ex})) \rangle$ with with $(O',\{e_1,\ldots,e_n\}) = f^\mi{extract}(\mi{ex},L)$ and $\pi_\mi{ct}(e_1) \leq \cdots \leq \allowbreak \pi_\mi{ct}(e_n)$.
\end{definition}

We depict a sequential encoding of the running example in \autoref{fig:encoding} b).
The events for process executions are ordered according to the complete timestamp of the event. The resulting sequence is attributed with the different feature values for each event. 
This encoding respects the timely order of events. 
\textit{However, it does not respect the true precedence constraints of the event log}: By merging all events into one sequence, some event pairs are forced into a precedence relationships they did not exhibit in the event log (cf. \autoref{sec:introduction}).
A graph encoding of features may be used for extensive visualization, applying graph algorithms or for utilizing graph neural networks~\cite{GNN}.
\begin{definition}[Graph Encoding]
Let $L {=} (E,O,OT,\pi_\mi{ct},\pi_\mi{st},\pi_\mi{trace},\pi_\mi{act},\allowbreak \pi_\mi{att})$ be an event log $\mi{EX}\subseteq  \mi{consub}(\mi{OG}_L)$ be a set of extracted process executions. Let $F_L = \{f_{L,1}, \ldots, f_{L,m}\} \subseteq E \times \mi{EX} \nrightarrow \mathbb{R}$ be a set of features. 
For an extracted execution $\mi{ex} \in \mi{EX}$, the graph of the corresponding process execution $p=(O',E') = f^\mi{extract}(\mi{ex},L)$ is defined by $G_p = (E',K) $ with edges $K = \{(e,e') \in E' \times E' \mid e \neq e' \wedge o \in O' \wedge 
\langle e_1,\ldots, e_n\rangle \in \pi_\mi{trace}(o) \wedge e = e_i \wedge e' = e_{i+1} \wedge i \in \{1,\ldots, n-1\} \})$. The graph encoding is defined by $G_\mi{feat}(p,F_L) = (E',K,l)$ with a node labeling function
$l(e) = \{f_L(e, \mi{ex}) \mid f_L \in F_L\}$ for any $e \in E'$.
\end{definition}

An example of the graph-based feature encoding for our running example is depicted in \autoref{fig:encoding} c). Each process execution is associated with a graph. Each node of the graph represents the feature values of an event.
\section{Use Cases}
\label{sec:casestudies}
\begin{table}[t]
    \centering
    \caption{Results for the different models based on different encodings.}
    \label{tab:results}
    \resizebox{5.5cm}{!}{
    \begin{tabular}{|c|c|c|c|}\hline
         & \textbf{Regression} & \textbf{LSTM} & \textbf{GNN}  \\\hline
        \textbf{Baseline MAE} & \multicolumn{3}{c|}{$0.7598$} \\\hline
        \textbf{Train MAE} & $0.5101$ & $0.4717$ & $0.4460$ \\\hline
        \textbf{Validation MAE}&NA&$0.4625$&$0.4534$\\\hline
        \textbf{Test MAE} & $0.5087$ & $0.4568$ & $0.4497$\\\hline
    \end{tabular}}
\end{table}
In this section, we evaluate our framework by providing six use cases. We pursue two evaluation goals with this approach: First, we aim to showcase the generalizability of the framework by providing a collection of common process mining tasks the framework can be applied to. Second, we aim to showcase the feature's and encoding's effectiveness in the use cases. Over the last years, explainable AI has been increasingly employed to make predictive process monitoring transparent~\cite{GalantiExplainablePredictiveMonitroing,HuangCounterfactualExplanation}. Through the use of \textsc{SHAP}~\cite{shap} values, we are able to quantify feature importance as well as structural importance of sequential and graph-based encoding.

The use cases are split into two parts: three visualization and three prediction use cases. 
We use a real-life loan application event log~\cite{vanDongen2017} as an OCEL. An event can be related to an application and multiple loan offers as objects.
We use tabular, sequential, and graph-based encoding to gain insights into the process through the visualization use cases. The prediction use cases aim at predicting the remaining time of an event's process execution (\textbf{P3}) using three different techniques for the different encodings: regression (tabular), LSTM neural networks (sequential), and GNNs (graph-based).
We use the same features for each encoding: Preceding activities (\textbf{C2}), average previous requested amount (\textbf{D1}), the elapsed time (\textbf{P2}), and the previous number of offers (\textbf{O3}). We use a 0.7/0.3 train/test split of the same events for each model for comparability reasons. We set aside 20\% of the training set as a validation set. The performance is assessed using the \emph{Mean Absolute Error (MAE)} of the normalized target variable.  Furthermore, we provide a baseline MAE achieved by predicting the average remaining time of the training set. The summarized results are depicted in \autoref{tab:results}.

We provide an open-source python implementation of our framework\footnote{\url{https://github.com/ocpm/ocpa}}. 
The framework can be extended with new features and adapted algorithms. 
{
\begin{figure}[t]
    \centering
    \includegraphics[width=0.75\columnwidth]{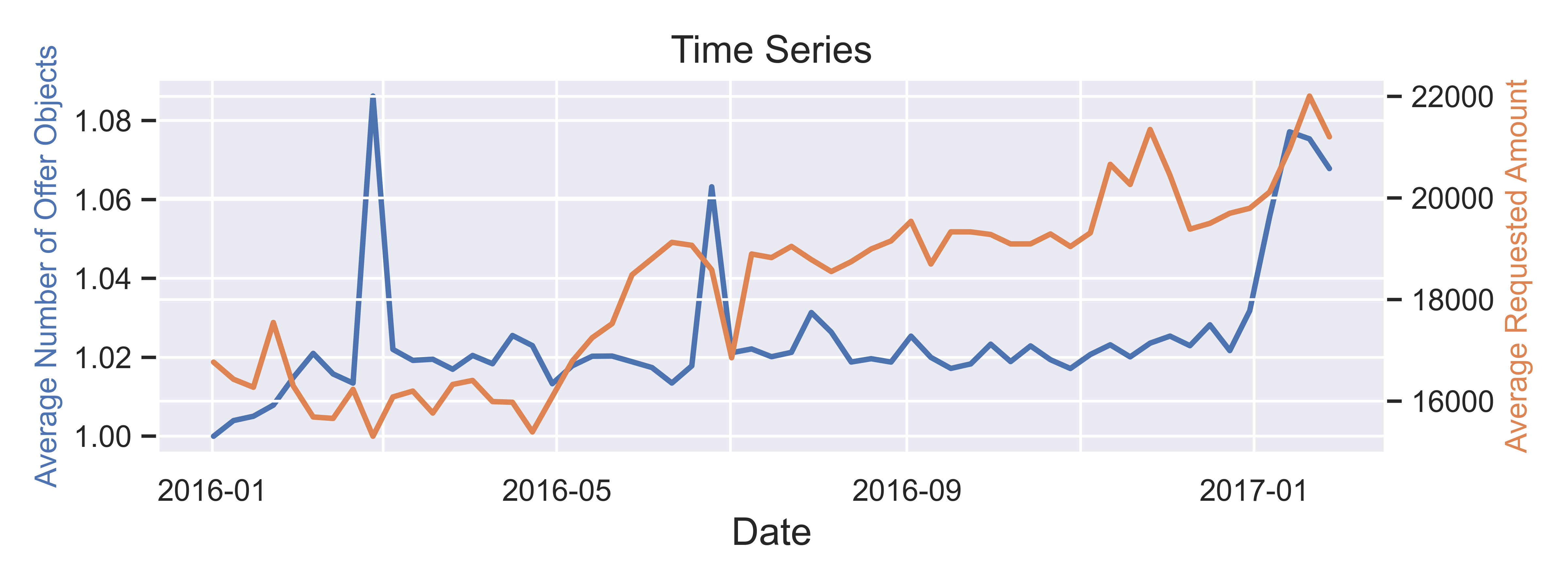}
    \caption{Time series describing two features over time: the weekly average number of loan offers per event and the weekly average requested amount for each application. Using this evaluation some initial insights can be generated, e.g., the gradual increase in requested amount over time.}
    \label{fig:timeseries}
\end{figure}}

\subsection{Tabular Encoding}
\paragraph{Visualization}
We split the event log into subsequent sublogs containing the events of one week each. For each sublog, we extract the average requested amount (\textbf{D3}) and the number of offers per event (\textbf{O6}). The resulting time series is depicted in \autoref{fig:timeseries}.
We can observe the dynamics of the process over time, e.g., the increase in the requested amount over time. Furthermore, we can observe that the number of offers is stable, except for a few short spikes.

\paragraph{Prediction}
We use a linear regression model to predict the remaining time based on the tabular encoding (cf. \autoref{tab:results}). This is an object-centric adaption of use cases \cite{DBLP:conf/otm/DongenCA08,DBLP:conf/otm/FolinoGP12}.
We generate the \textsc{SHAP} values, i.e., the impact of different features on the individual model prediction, for 1000 predictions of the test set. The resulting bee swarm plot is depicted in \autoref{fig:beeswarm} (left side). Red points indicate a high feature value. The more they are positioned to the left, the more the feature value reduces the model's prediction. Therefore, the combination of color and position gives insights into the feature value's impact on the model output. We can, e.g., observe a high decreasing impact of the existence of the \textit{Call} activity in the preceding activities to the predicted remaining time. 
One can also observe an impact of the new object-centric feature of the number of previous objects of type offer on the predicted remaining time: the more offers were previously recorded in a case, the lower the predicted remaining time. 
In conclusion, the selected set of object-centric feature adaptations yields valuable information for a predictive model. 

{
\begin{figure}[t]
    \centering
    \includegraphics[width =\columnwidth]{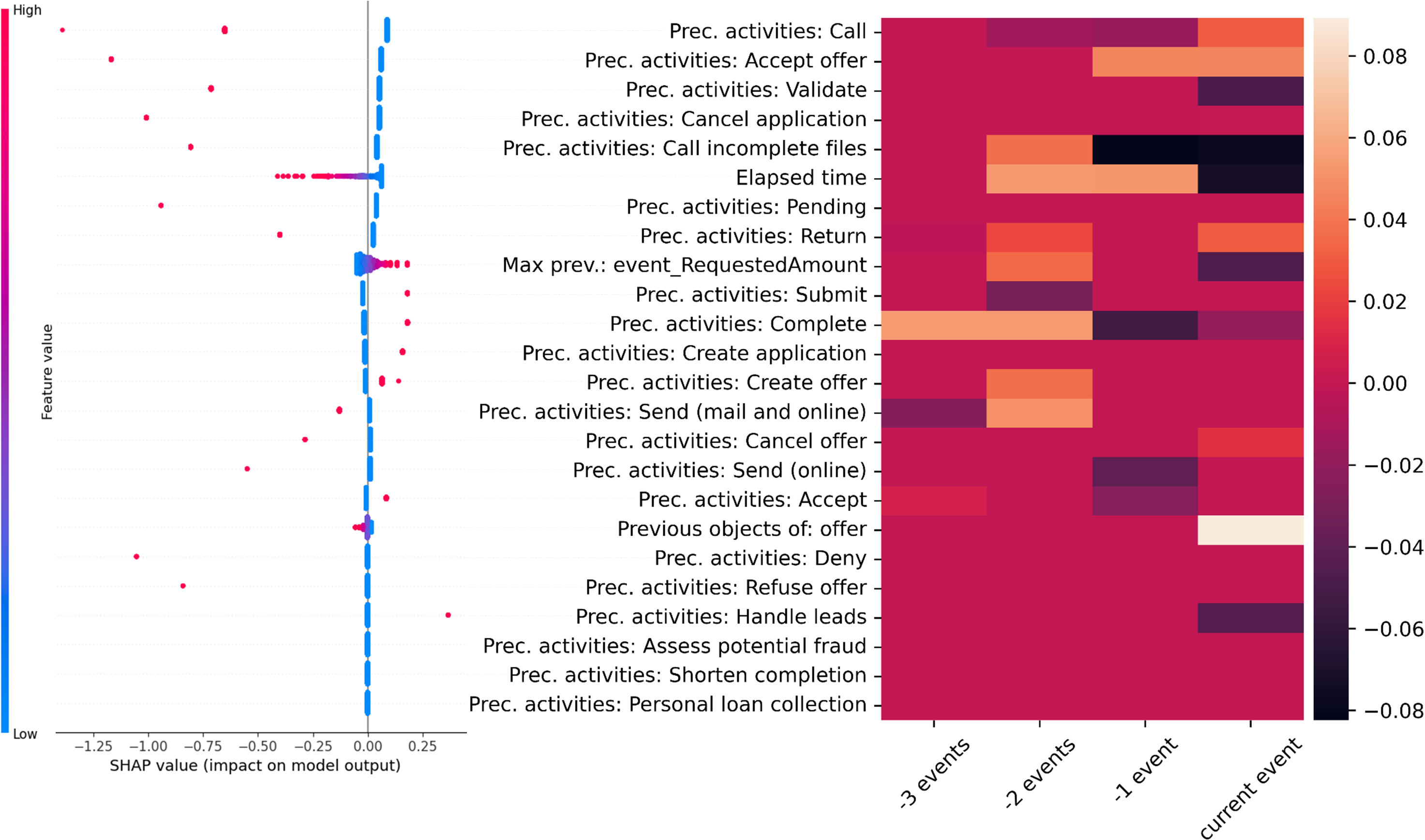}
    \caption{Left: Bee swarm plot of \textsc{SHAP} values for the regression model, showing the aggregated importance of different features to the predictions. Right: \textsc{SHAP} values of one LSTM prediction, visualized for the different positions of the input sequence. }
    \label{fig:beeswarm}
\end{figure}}
{
\begin{figure}[t]
    \centering
    \includegraphics[width=0.96\columnwidth]{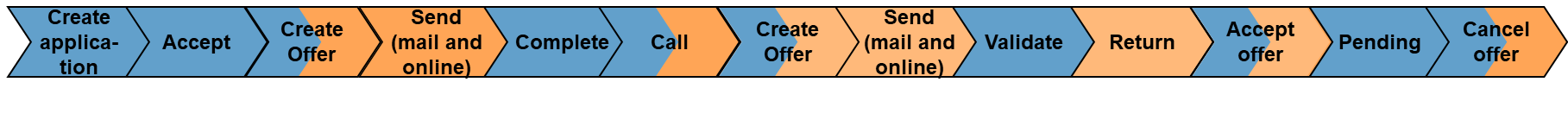}
    \caption{Sequential variant visualization of a process execution enriched with the object information (blue = application, orange = first and second offer).}
    \label{fig:sequence_var}
\end{figure}}
\subsection{Sequential Encoding}
\paragraph{Visualization}
We choose one specific process execution and extract the sequential encoding for the current event's activity (\textbf{C5}) and the event's objects (\textbf{O4}) features. The result is a variant enriched by object information, depicted in \autoref{fig:sequence_var}. Even though such a visualization might have misleading causality information for events between objects, one can already retrieve some valuable insight into the intra-object order and the overall activities of an execution.

\paragraph{Prediction}
We use a neural network with two 10-hidden-node LSTM layers to predict the remaining time of the sequentially encoded features. We use subsequences of length four (cf. \autoref{tab:results} for results). This is an object-centric adaption of use cases~\cite{LSTMPrediction,DBLP:conf/bpm/LeontjevaCFDM15}.
The regression use case already covered the importance of features for the prediction. Therefore, we focus on the importance of the sequential encoding in this use case. We use \textsc{SHAP} values for each feature of the four positions in the sequential encoding. The calculated feature impacts for an individual prediction are depicted in \autoref{fig:beeswarm} (right side). The more the value diverges from zero, the higher the feature's impact on the model's output. We observe features with high importance among all four positions of the sequence. Therefore, the model utilizes the sequential encoding of the features, showcasing its usefulness.

\subsection{Graph Encoding}

\paragraph{Visualization}
\autoref{fig:graphCS} a) depicts the graph-based variant visualization retrieved from OC$\pi$~\cite{OCpi} of the same process execution as \autoref{fig:sequence_var}. Using the graph, one can place concurrent events in two different lanes according to their objects, not indicating any precedence between them. One can intuitively determine the concurrent paths in the variant and the interaction of different objects. For large process executions, this provides structured access to the control-flow of the underlying process.

\paragraph{Prediction}
We use the graph-based feature encoding as an input for a GNN. The GNN contains two graph convolution layers. Each node in both layers has a size of 24. Input graphs are constrained to four nodes (cf. LSTM use case). We read the graphs out by averaging over the convoluted values, summarizing to one predicted remaining time (cf. \autoref{tab:results} for results). This is an object-centric adaptation of use cases~\cite{9004296,9533742}.
We adapt \textsc{SHAP} values to determine the importance of graph edges to the predicted remaining time. \autoref{fig:graphCS} b) depicts the calculated values for one graph instance. The more the value of an edge diverges from zero, the higher its existence impacts the model's prediction. We observe substantially different values for all edges: While some edges have a relatively low negative or positive impact on the model's output, the presence of other edges heavily impacts the predicted remaining time. Therefore, the graph structure itself yields important information for predicting the remaining time. 
{
\begin{figure}[t]
    \centering
    \includegraphics[width=0.99\columnwidth]{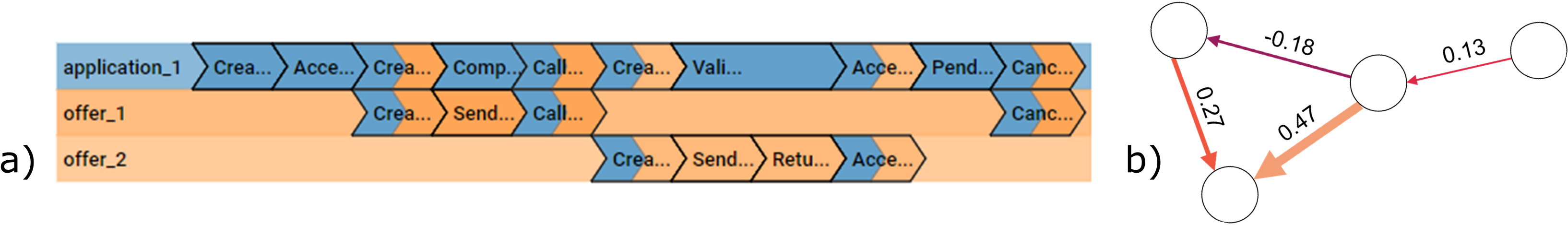}
    \caption{Use cases for the graph encoding: a) activities and objects of one process execution. Shared events between objects are colored with multiple colors. b) shows the importance of different edges of one instance graph when predicting.}
    \label{fig:graphCS}
\end{figure}}
\section{Conclusion}
\label{sec:conclusion}
We introduced a general framework to extract and encode features from OCELs. Currently, object-centric event data needs to be flattened to apply process enhancement techniques to the data. This leads to inaccurate features. Additionally, no feature encoding is available to express the graph-like structure of object-centric event data. Our framework calculates features natively on the object-centric event data, leading to accurate features. Furthermore, we provide a graph-based encoding of the features, preserving the underlying structure of an OCEL. We show the utility of the features and encodings in six use cases, a visualization and prediction use case for each of the three encodings. This framework lays a foundation for future machine learning approaches utilizing object-centric event data and new algorithms using our encodings as a basis.

We provide a collection of use cases showing the applicability of our framework for extracting and encoding features. 
For each of our framework steps, interesting future research directions are present: Which feature work well with which encoding? What are the best prediction techniques for which encoding? How to optimize existing network architectures to achieve maximum results?
Furthermore, investigations of new features derived from the graph structure and object-multiplicity as well as further traditional features not included in de Leoni et al.'s framework~\cite{GeneralFrameworkCorrelating} is an interesting direction for future research.

%
%
%
%
\bibliographystyle{splncs04}
\bibliography{bibliography}

\end{document}